\documentclass[letterpaper, 10 pt, conference]{ieeeconf}  % Comment this line out if you need a4paper

\usepackage{cite}
\usepackage{amsmath,amssymb,amsfonts}
\usepackage{subcaption}
\usepackage{algorithmic}
\usepackage{graphicx}
\usepackage{textcomp}
\usepackage{xcolor}
\usepackage{cleveref}

\IEEEoverridecommandlockouts                              % This command is only needed if 
\title{\LARGE \bf
Robust Flower Cluster Matching Using The Unscented Transform
}

\author{Andy Chu$^{1}$, Rashik Shrestha$^{2}$, Yu Gu$^{1}$, Jason N. Gross$^{1}$% <-this % stops a space
\thanks{This study was supported in part by USDA NIFA Award 2022-67021-36124 "Collaborative Research: NRI: StickBug - an Effective Co-Robot for Precision Pollination".}% <-this % stops a space
\thanks{$^{1}$Department of Mechanical, Materials, and Aerospace Engineering, West Virginia University, Morgantown, USA }%
\thanks{$^{2}$Lane Department of Computer Science and Electrical Engineering, West Virginia University, Morgantown, USA}%
}

\begin{document}

\maketitle
\thispagestyle{empty}
\pagestyle{empty}

%%%%%%%%%%%%%%%%%%%%%%%%%%%%%%%%%%%%%%%%%%%%%%%%%%%%%%%%%%%%%%%%%%%%%%%%%%%%%%%%
\begin{abstract}
Monitoring flowers over time is essential for precision robotic pollination in agriculture. To accomplish this, a continuous spatial-temporal observation of plant growth can be done using stationary RGB-D cameras. However, image registration becomes a serious challenge due to changes in the visual appearance of the plant caused by the pollination process and occlusions from growth and camera angles. Plants flower in a manner that produces distinct clusters on branches. This paper presents a method for matching flower clusters using descriptors generated from RGB-D data and considers allowing for spatial uncertainty within the cluster. The proposed approach leverages the Unscented Transform to efficiently estimate plant descriptor uncertainty tolerances, enabling a robust image-registration process despite temporal changes. The Unscented Transform is used to handle the nonlinear transformations by propagating the uncertainty of flower positions to determine the variations in the descriptor domain. A Monte Carlo simulation is used to validate the Unscented Transform results, confirming our method's effectiveness for flower cluster matching. Therefore, it can facilitate improved robotics pollination in dynamic environments.
\end{abstract}
\section{Introduction}
\label{sec:introduction}

Although global agriculture relies heavily on pollination, evidence has shown that the population of natural pollinators is decreasing, raising concerns about food and the economy \cite{potts2016assessment}. In response to these concerns, advancements in automation in agriculture allow farmers to increase crop yields, reduce labor costs, and benefit from diversifying farming systems \cite{lowenberg2020economics}. The development of these systems is often developed in more structured and controlled environments, such as greenhouses \cite{edan2009automation}. Automation requires data collection and feedback to overcome challenges associated with agricultural tasks in precision robotics. Continuous monitoring of crops and the registration of multiple sensor data over time are necessary for computer vision to accurately track plant development and health. In precision operations such as \cite{ohi2018design,smith2024design,droukas2023survey}, the detection of flower centers as targets is needed to assist in the agricultural task of pollination. Computer vision in robotics often uses stereo vision for these precision robotic agriculture tasks, which require the use of the robot arm to interact with the target \cite{wang20173d}.

\begin{figure}[htb]
    \centering
    \includegraphics[width=.75\linewidth]{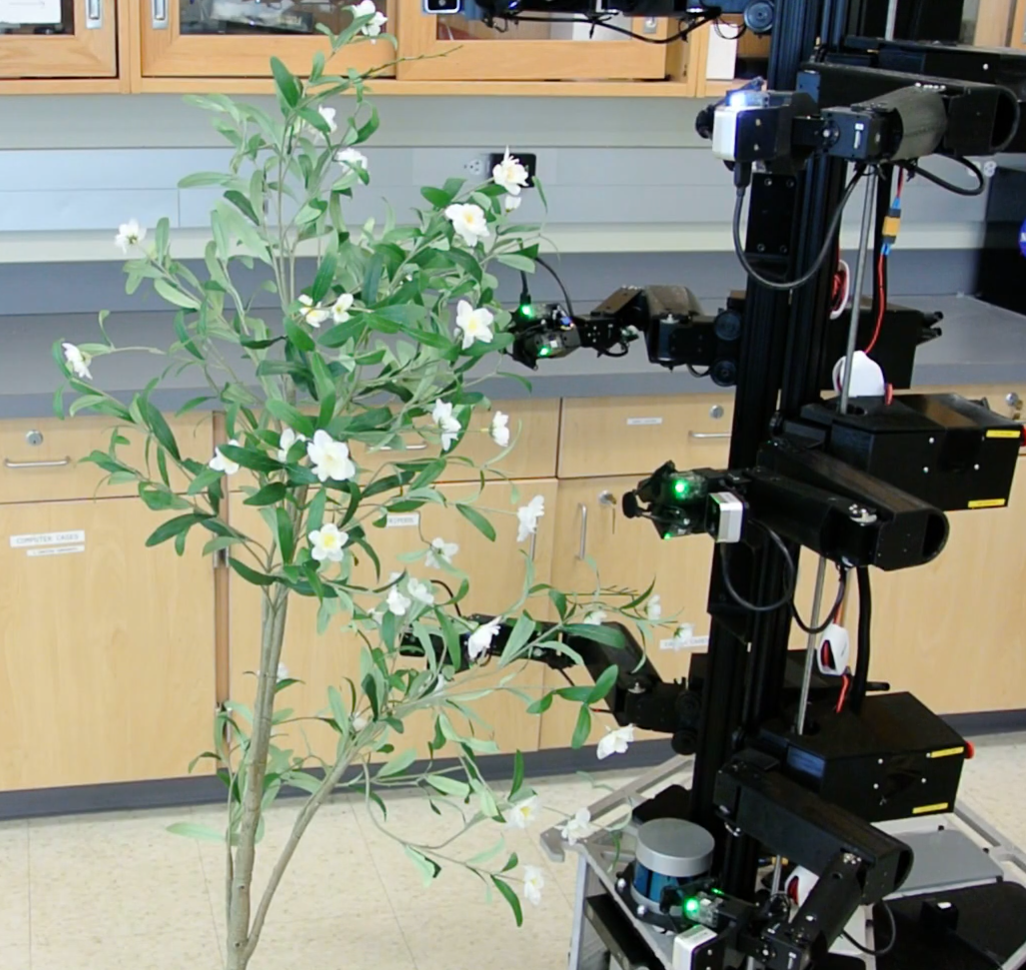}
    \caption{Flower clusters on an artificial plant being pollinated by Stickbug, a six-armed robot pollinator (WVU Photo).}
    \label{fig:intro}
\vspace{-10pt} \end{figure}

There are challenges to flower detection, such as camera view obstruction and crop occlusions caused by crop manipulation \cite{chandel2015occlusion}. Manipulation of branches for operations causes the branch to sway, resulting in movement of the targeted points. This is shown by simulating the physical properties of a branch of a tree for manipulation use \cite{jacob2024learning}. This results in a visual change, making the registration of the flower cluster harder, in addition to the already existing sensor noise. Spatial information can be simplified to a descriptor that is useful in defining the cluster.

In this paper, the challenges of flower cluster re-identification are addressed. Plant growth and interactions with robot manipulators introduce visual changes captured by vision sensors. Our approach simplifies flower clusters into a two-dimensional descriptor and employs the Unscented Transform to propagate descriptor uncertainty. To validate this method, Monte Carlo is applicable in the estimation of nonlinear statistical data \cite{james1980monte,metropolis1949monte}. Although often used as a tolerance estimation for mechanical design, the theoretical method can also be applied to any statistical analysis. The formulation of Monte Carlo relies on the Law of Large Numbers, and the convergence accuracy depends on the size of the random variables. For matching, the Mahalanobis distance is used alongside the chi-squared distribution to detect outliers \cite{gu2016fault}. 

The main contribution of this research is a novel method for generating cluster descriptors and using the Unscented Transform for feasible real-time flower cluster matching, despite plant growth and structural changes, validated using a Monte Carlo simulation. It addresses descriptor nonlinearity and estimates flower locations, which can be applied in precision agriculture tasks like harvesting, pollination, and growth monitoring, and extends to landmark-based localization. An example of such applications is shown in Fig. \ref{fig:intro}. 
\section{Related Works}

% cite 2 papers one aerial and ground
In the field of precision agriculture, a multitude of studies have investigated the monitoring of crop fields. These methods employ the use of aerial vehicles and ground vehicles. UAVs offer high-resolution imagery, enabling the extraction of visual data necessary to assess crop health through computer vision and artificial intelligence. The use of multi-rotor and fixed-wing UAVs facilitates faster and more cost-effective monitoring over large areas, providing higher spatial and temporal resolution compared to satellite imagery \cite{cuaran2021crop}. UGVs and mobile robots are often found in fields. Robotic systems are often tested for agricultural harvesting applications when equipped with RGB-D cameras and computer vision systems \cite{oliveira2020agricultural}. Technological advancements assist the development of these methods in the field of computer science, which allow the detection and classification of objects.

% Descriptors
The challenge of data association across multiple images has been in multiple studies. The applications of these techniques tend to address the global view of crops. While several computer vision methods like SIFT or ORB can be used to identify flower clusters, there are challenges due to similar features and appearances, occlusions, lighting, and variability due to growth. The works of \cite{chebrolu2018robust} proposed a novel method that addresses registration issues under large visual changes by exploiting geometrical information. Geometric information is used to create a descriptor that is scale-invariant and uniquely describes the set of points. In the work of \cite{liu2022farmland}, an optimized SIFT algorithm is used to stitch high-resolution farmland aerial images by feature points and filtering the mismatches. These methods allow faster and more reliable agricultural monitoring.

The work of \cite{rhudy2013understanding} addresses the high-level language of the earlier literature \cite{julier1997new,van2004sigma} and simplifies the literature to make the use of the Unscented Transform accessible. The Unscented Transform approximates a Gaussian distribution after a nonlinear Transform. In the context of an Unscented Transform, there is a lot of existing literature on non-linear state estimation. Often, the Unscented Transform is used in the form of the Unscented Kalman Filter (UKF), which was developed to achieve better results compared to the Extended Kalman Filter (EKF) \cite{wan2000unscented}. In a case study for attitude estimation, it was found that UKF was more robust in terms of bias and scale factors of the Inertial Measurement Unit (IMU) compared to the EKF \cite{rhudy2013sensitivity}.

\section{Methodology}

\begin{figure}[htb!]
    \centering
    \includegraphics[width=6cm,height=4.5cm]{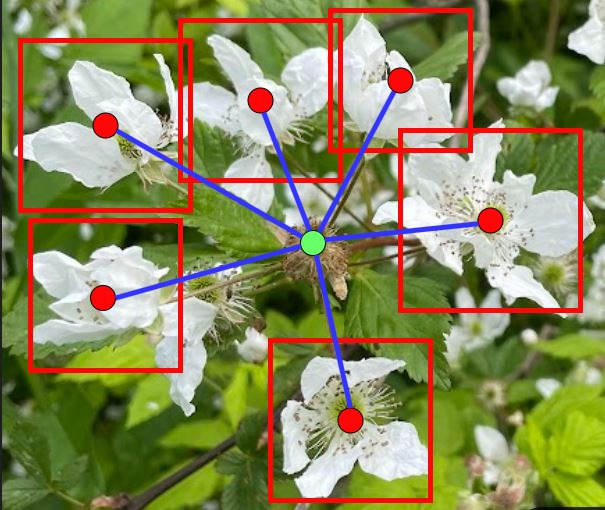}
    \caption{A cluster consisting of 6 bramble flowers and its relation to the centroid used for creating the descriptor used to describe its geometrical information (WVU Photo).}
    \label{fig:descriptor}
\vspace{-10pt} \end{figure}

\subsection{Objective Statement}
The objective of this study is to develop a flower cluster descriptor that accounts for variations due to growth and manipulation. By leveraging the Unscented Transform, the proposed approach enhances the consistency of cluster re-identification. Additionally, an evaluation metric is established to determine the accuracy and reliability of the descriptors.

\subsection{Assumptions and Overview}
This section describes the approach used to simulate flower clusters, design the Unscented Transform, and establish the evaluation metric. The simulation process models realistic possible states of the flower due to the growth and manipulation of its branches. The Unscented Transform is designed to handle the nonlinear transform of the cluster descriptors. Finally, an evaluation metric is established to determine the accuracy and robustness. The following assumptions were made regarding simulation, testing, and data collection:

\begin{itemize}
    \item The number of flowers in a cluster ranges from 3 to 6.
    \item The plant has displacements due to growth and manipulation that are reasonable (i.e., displacements between 0.01 to 0.05 meters)
    \item The flowers in the cluster remain detected after growth and manipulation
    \item An artificial bramble plant with a single cluster with manual manipulation to simulate movement and growth is evaluated.
\end{itemize}

The development of the Monte Carlo simulation was necessary to validate the use of the Unscented Transform to propagate the nonlinearity of the descriptor. Monte Carlo Simulation produces descriptors of multiple random variations of an initial position to simulate the uncertainty of flower locations in the cluster. By sampling the sigma points, the Unscented Transform captures the nonlinear transform of the positions to descriptors accounting for the propagation of nonlinear descriptor function. The covariances of both Monte Carlo and Unscented Transform are then evaluated using Frobenius Norm. The effectiveness of the matching is determined by evaluating the Mahalanobis Distance to its chi-square distribution.

% Write about the parameters for UT and MC

\subsection{Extracting Position And Calculating Descriptor}

The setup includes a robotic manipulator equipped with an eye-in-hand configuration. The manipulator arm performs pollination by gently rubbing its end-effector against the flower. This process is repeated multiple times, during which positional information of the flower cluster is collected.

The sequence of $N$ images is captured as $\mathbf{I}=\{I_0, I_1,..., I_N | I_i\in\mathbb{R}^{H \times W \times 3}\}$. Corresponding depth images $\mathbf{D}=\{D_0, D_1,..., D_N | D_i\in\mathbb{R}^{H \times W} \}$ are gathered, along with camera poses $\mathbf{P}=\{P_0, P_1,..., P_N | P_i\in\mathbb{R}^{4 \times 4}\}$ representing 3D transformation matrices.

The Grounding DINO~\cite{liu2025grounding} model $\mathcal{D}$ for detecting the pixel coordinates of the flowers $u_i$ within the image frame $I_i$, conditioned upon a text prompt $t$.

\begin{equation}
u_i = \mathcal{D}(I_i, t), u_i \in \mathbb{R}^{F_i \times 2}
\end{equation}

where $F_i$ denotes the number of flowers detected in the given sequence frame $i$. The camera intrinsic parameters $K$ and depth image $D_i$ lift the 2D pixel coordinates $u_i$ to 3D coordinates $x_i$.

\begin{equation}
x_i = D_i[u_i]\frac{K^{-1}\tilde{u}_i}{||K^{-1}\tilde{u}_i||}
\end{equation}

where, $\tilde{u}_i$ represent homogeneous form of $u_i$, $D_i[u_i]$ is depth values extracted at the pixel coordinates $u_i$, and $||.||$ represents L2 norm. Note that $D_i[u_i]$ is the depth along the ray (not along the z-axis of the camera coordinate frame), necessitating the normalization factor.

Finally, the 3D flower coordinates are transformed into the world coordinate frame. This is necessary to ensure a consistent flower position across the frames while the camera moves.

\begin{equation}
x^W_i = P_i.x_i
\end{equation}

where $x^W_i$ represents 3D flower coordinates in the world coordinate frame.

The flower coordinates in the cluster are then used to generate the descriptors as seen in Fig. \ref{fig:descriptor}. Descriptors are designed to be less sensitive to the changes in the growth of the plants and additional flowers. To achieve this, the descriptor is based on the spread of data relative to the centroid of the cluster. The first descriptor is inertia, and the second descriptor is the average distance of the flowers' centers to the cluster's centroid \cite{edwards1965method}.

\begin{equation}
    \text{inertia} = \sum_{i=1}^{N} \|\mathbf{pos}_i - \mathbf{c}\|^2
\end{equation}

\begin{equation}
    \text{average distance} = \frac{1}{N} \sum_{i=1}^{N} \|\mathbf{pos}_i - \mathbf{c}\|
\end{equation}

In these equations, $N$ is the number of flowers, $i$ is the index of the flower, $\mathbf{pos}$  denotes the positions of the bounding box centers of the detected flowers, and $\mathbf{c}$ is the centroid of the cluster. By relating the descriptor to the centroid of the cluster, sensitivity is reduced.

\subsection{Monte Carlo Simulation}
% Explain MC and cite the original paper
% Explain my Monte Carlo

The purpose of a Monte Carlo is to help analyze how the noise on the positioning of each flower affects the computed descriptors of the flower positions.  The simulation parameters are defined by the number of flowers in the cluster, the expected camera noise, and the displacement of flower center positions due to the detection process. In this case, the setup includes three flowers with an initial noise level of 0.01 meters, chosen to reflect real collected data. To ensure a robust analysis for Monte Carlo to be good, a larger number of samples is needed, therefore, the number of clusters that were to be simulated was 10,000 iterations. To better understand the impact of noise, the analysis was extended to cover a range of 0.01 to 0.05 meters, representing realistic displacement caused by plant movement, pollination, and natural growth.

A fixed seed is used to maintain reproducibility. Utilizing Matlab's rand function, the initial 3D positions of points representing the flower center are generated about the world frame with uniform distribution. Gaussian noise is added to the generated initial positions by using Matlab's randn function to simulate the uncertainty in the measurement of the positions. The perturbed positions are then processed through a descriptor function based on the centroid, converting them into a descriptor including the inertia and average distance of points to the centroid. The descriptors of the perturbed positions are then logged and evaluated. The covariance matrix and mean of the computed descriptors are calculated, providing insight into the effects of the perturbations of the descriptors.

\subsection{Unscented Transform}

In the context of this work, the Unscented Transform is used to handle the nonlinear transformations inherent in descriptor calculations. The UT propagates sigma points through a non-linear function, and the selection of sigma points is controlled by scaling parameters\cite{julier1997new, rhudy2013understanding, wan2001unscented,van2004sigma}:
\begin{itemize}
    \item $\alpha$: Controls the spread of the sigma points, typically ranging from $10^{-4}$ to 1. Smaller values result in a tighter distribution, while larger values produce a wider spread.
    \item $\beta$: Incorporates prior knowledge of the distribution. For a Gaussian prior, $\beta = 2$ is optimal.
    \item $\kappa$: Typically set to 0, as recommended in \cite{julier1997new}.
    \item $\lambda$: A scaling parameter derived from $\alpha$ and $\kappa$, which influences the sigma points' placement.
    \item $\eta^m$ and $\eta^c$: Weights used to compute the transformed mean and covariance, respectively.
    \item $L$: The length of the state vector
\end{itemize}

\begin{equation}
    \lambda = \alpha^2(L+\kappa)-L
\end{equation}
\begin{equation}
    \eta_0^m = \lambda/(L+\lambda)
\end{equation}
\begin{equation}
    \eta_0^c = \lambda/(L+\lambda)+1-\alpha^2+\beta
\end{equation}
\begin{equation}
    \eta_i^m = \eta_i^c = 1/[2(L+\lambda)], \hspace{10pt} i=1,...,2L
\end{equation}

The sigma point matrix $\chi$ is an $L(2L+1)$ matrix of sigma points, where each column represents a sigma point:

\begin{equation}
    \chi = [\bar{x}, \bar{x}+\sqrt{L+\lambda}\sqrt{P_x}, \bar{x}-\sqrt{L+\lambda}\sqrt{P_x}]
\end{equation}

The Unscented Transform initializes the positions of the 3D points as its prior mean, denoted as $\bar{x}$, and then the covariance, denoted as $P_x$, and the noise variance. The square root operation of $P_x$ is performed using the Cholesky decomposition. Sigma points are then generated and propagated through the nonlinear descriptor function. Then, the mean and the covariance of the propagated sigma points can be computed.

\subsection{Evaluation Methods and Determination of Matching}
\begin{figure*}
    \centering
    \includegraphics[width=.9\linewidth]{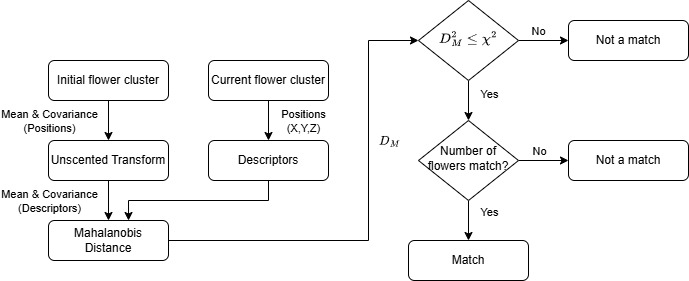}
    \caption{The flower cluster matching utilizing the Unscented Transform pipeline}
    \label{fig:matching}
\end{figure*}

In this section, the various methods of evaluation used to validate the use of the Unscented Transform are presented. These methods include the use of the Frobenius Norm, Mahalanobis distance, Chi-Square Distribution, and an additional condition padding for matching criteria.

Frobenius Norm is used as a metric to determine the differences between two matrices. In the context of using the Monte Carlo to validate the Unscented Transform, the Frobenius Norm is used as a method to evaluate the similarity of the two covariance matrices of the descriptor.  This is done by taking the square root of the difference, denoted as 
 ${a_{ij}}$ between the two covariance matrices, resulting in a single numerical value representing the matrix difference in terms of a magnitude denoted by $\mathrm{||A||_F}$. The Frobenius norm of a Matrix A is defined as:
 
\begin{equation}
    \| A \|_F = \sqrt{\sum_{i=1}^{m}\sum_{j=1}^{n} |a_{ij}|^2}
    \label{eq:frobenius_norm}
\end{equation}

Where $a_{ij}$ represents the elements of Matrix A, and $m$ and $n$ denote the number of rows and columns, respectively.

The Mahalanobis distance determines the correlation of a point to a distribution. The Mahalanobis distance measures the relative distances to the overall mean of the multivariate descriptors. The equation is defined with $D_M$ representing the Mahalanobis distance, $x$ as the descriptor vector, $y$ as the mean vector of the Unscented Transform, $\mathbf{S}^{-1}$ as the inverse of the covariance matrix, and $T$ as the transpose.

\begin{equation}
    D_M(\mathbf{x}, \mathbf{y}) = \sqrt{(\mathbf{x} - \mathbf{y})^T \mathbf{S}^{-1} (\mathbf{x} - \mathbf{y})}
    \label{eq:mahalanobis_distance}
\end{equation}

The Mahalanobis distance was selected to define a threshold for multidimensional variables by measuring their "closeness" to the overall mean of the multivariate descriptors. It follows a chi-square distribution with $n$ degrees of freedom and is used to identify outliers based on a predefined threshold. This threshold, determined by the inverse chi-square function, assesses whether a descriptor point belongs to the multivariate normal distribution at a given confidence level $\alpha$, where $d$ represents the degrees of freedom.

\begin{equation}
    \chi^{2}_{inv}(\alpha,d)
\label{eq:chi_square_inverse_threshold}
\end{equation}

If ${D_M}^2>\chi^{2}_{inv}(\alpha,d)$, then the point would be considered an outlier for the given confidence level. 

% Explain padding was evaluated where it can help increase the number of matches
Padding was evaluated to determine its impact on the number of matches. Adding padding on the covariances along the diagonal elements increases the uncertainty of the state estimation. This leads to more matches to accommodate the higher variations in the data. However, on the contrary, it could lead to increased false detections.

Matching criteria were established to determine whether a flower cluster corresponds to a given descriptor. A cluster is considered a match if the Mahalanobis distance satisfies ${D_M}^2<\chi^{2}_{inv}(\alpha,d)$, ensuring that the descriptor is not classified as an outlier. To further enhance matching, an additional condition is imposed to verify that the number of flowers assigned to the descriptor, $n$, matches the number of flowers in the cluster being evaluated. If both criteria are met, then the cluster being evaluated will be considered a match. The constraints are designed to help maintain the robustness of the matching process. The matching entire pipeline is illustrated in Fig. \ref{fig:matching}.

\section{Experiments and Results}
\subsection{Experimental Setup and data collection}

The testing workspace features a Universal Robots UR5 Robot Arm Manipulator with an end-effector mounted RealSense camera, specifically the D405 and D435 models. Data collection was completed using a PC (12-core CPU, 32GB of RAM, and Linux 22.04) while the pose estimation was completed on a server with an RTX 3070 Ti GPU and 64GB of RAM.

To extend testing beyond the natural flowering season of bramble plants, an artificial plant was constructed. Four different datasets of flower clusters were collected and pruned to remove unreliable detections. Perturbations were manually introduced by shaking the plant. While data generation isn't standardized and reproducible per trial, it serves as a practical substitute for the plant movements due to external forces and growth.

Simulations were performed on a Windows 11 Pro PC (AMD Ryzen 7 5800H, 8 cores, 16GB RAM) using MATLAB and the Statistics and Machine Learning Toolbox.

%% NOTE: Make a table for sets of data
\subsection{Evaluation Results}

\begin{table}[h]
\caption{Observation of noise effect (0.01 to 0.05) on Unscented Transform with a cluster of 3 flowers}
\label{tab:noise_effect}
\begin{center}
\renewcommand{\arraystretch}{1.25} % Increases row height
\begin{tabular}{|c|c|c|}
            \hline
            \textbf{Noise} & \textbf{Frobenius Norm} & \textbf{Percentage Outlier} \\
            \hline
            0.01 & \(1.2314 \times 10^{-6}\)  & 5.01 \\
            \hline
            0.02 & \(4.1397 \times 10^{-6}\)  & 4.93\\
            \hline
            0.03 & \(4.1437 \times 10^{-5}\)  & 4.83\\
            \hline
            0.04 & \(1.5409 \times 10^{-4}\)  & 4.75\\
            \hline
            0.05 & \(4.0231 \times 10^{-4}\)  & 4.65\\
            \hline
\end{tabular}
\end{center}
\end{table}

Table \ref{tab:noise_effect} presents the impact of increasing noise levels (0.01 to 0.05 meters) on the Unscented Transform when applied to a cluster of three flowers. A total of 10,000 trials were conducted, where initial positions were perturbed within this range. The chi-square threshold was set at 5.9915 for a 95\% confidence level with two degrees of freedom. These perturbations were incorporated into the Unscented Transform for covariance estimation.

The covariance of the simulated descriptors was compared to that of the Unscented Transform using the Frobenius norm, revealing strong similarity, with a slight increase in difference as perturbations grew stronger. The effect of these perturbations on the number of outliers was also analyzed. The chi-square threshold remained constant at a 95\% confidence level for two dimensions. Results indicate that the Unscented Transform remains effective across the tested range, with higher noise levels causing a gradual decrease in outliers. It's important to note that the number of outliers is directly related to the set confidence level, as it determines the threshold.

\begin{table}[htbp]
    \centering
    \caption{Comparison of correct matches and false positives with and without 0.005 meter padding across 10,000 samples of clusters containing 3 to 6 flowers (conditioned on matching the number of flowers).}
    \renewcommand{\arraystretch}{1.25} % Increases row height
        \begin{tabular}{|c|c|c|c|}
            \hline
            \textbf{Padding (0.005)} & \textbf{Noise} & \textbf{Correct Match} & \textbf{False Positive (Avg)} \\
            \hline
            w/o Padding & 0.01 & 8820 & 736.1542 \\
            \hline
            w/ Padding & 0.015 & 9420 & 797.0514 \\
            \hline
        \end{tabular}
    \label{tab:padding_comparison}
\end{table}

Table \ref{tab:padding_comparison} presents the findings on the effect of padding on the covariance of the Unscented Transform. An initial noise level of 0.01 meters and a padding value of 0.005 meters were used. The objective of the analysis was to determine whether padding could improve the number of correct matches.

The simulation results indicate that padding is effective in increasing the number of correct matches. Specifically, adding 0.005 meters of padding to the covariance of the Unscented Transform increased correct matches to 600, while also introducing an average of 60.9 more false matches per correct match. This improvement comes with a trade-off where the padding increases correct matching, but it also increases the number of false matches.

% Real data
\subsection{Real Data Results}
\begin{figure}[htb!]
    \centering
    \begin{subfigure}{0.49\linewidth} % First subfigure
        \centering
        \includegraphics[width=\linewidth]{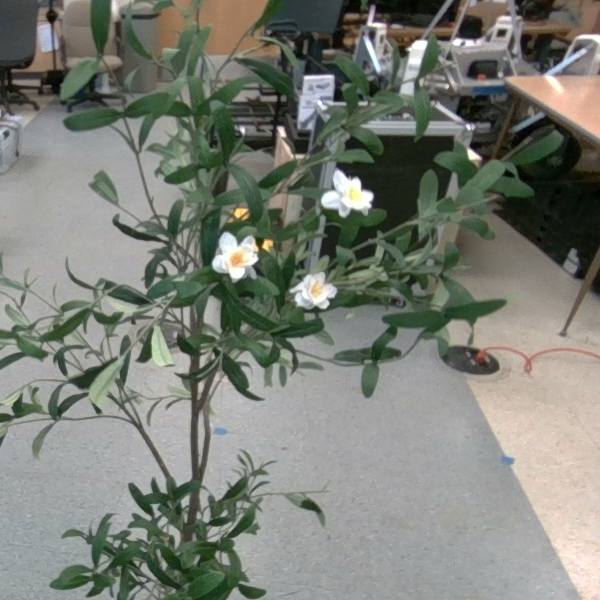}
        \caption{Dataset 1: A cluster of 3 flowers used for analysis.}
        \label{fig:subfig1}
    \end{subfigure}
    \hfill
    \begin{subfigure}{0.49\linewidth} % Second subfigure
        \centering
        \includegraphics[width=\linewidth]{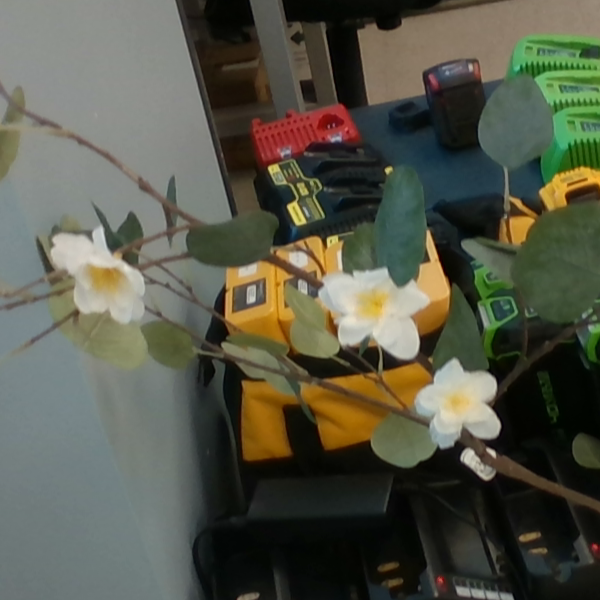}
        \caption{Dataset 2: A cluster of 3 flowers used for analysis.}
        \label{fig:subfig2}
    \end{subfigure}
    \vskip\baselineskip
    \begin{subfigure}{0.45\linewidth} % Third subfigure
        \centering
        \includegraphics[width=\linewidth]{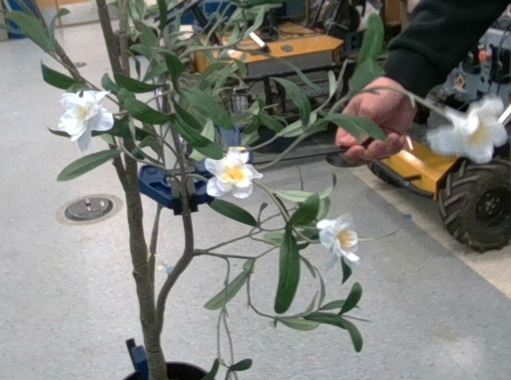}
        \caption{Dataset 3: A cluster of 4 flowers used for analysis.}
        \label{fig:subfig3}
    \end{subfigure}
    \hfill
    \begin{subfigure}{0.5\linewidth} % Fourth subfigure
        \centering
        \includegraphics[width=\linewidth]{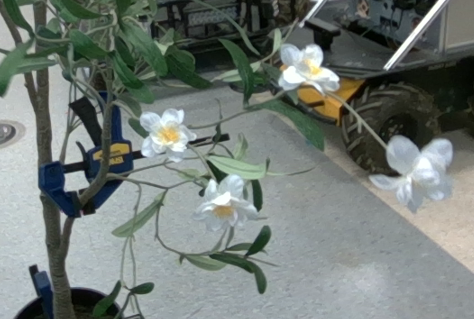}
        \caption{Dataset 4: A cluster of 4 flowers used for analysis.}
        \label{fig:subfig4}
    \end{subfigure}
    \caption{Flower clusters with 3 bramble flowers under moderate shaking (a–b) and 4 flowers under intense shaking (c–d), collected from an artificial plant for analysis (WVU Photo).}
    \label{fig:datasets}
\end{figure}

Four unique flower cluster configurations were designed and attached to the artificial bramble plant. A dataset of each flower cluster configuration was collected. Having two different cluster datasets allows the observation of the effectiveness of the Unscented Transform and its application in matching. The datasets are pruned before logging to ensure data quality. Specifically, only entries where the number of 3D positions matches the number of flowers are retained. Mismatches are discarded, and the data entry is removed. Fig. \ref{fig:datasets} presents the positions of the flowers in the respective Datasets. After pruning, Dataset 1 Fig. \ref{fig:subfig1} resulted in 4954 frames,  Dataset 2 Fig. \ref{fig:subfig2} resulted in 277 frames,  Dataset 3 Fig. \ref{fig:subfig3} resulted in 2437 frames, and Dataset 4 Fig. \ref{fig:subfig4} resulted in 1005 frames.

\begin{table}[htb]
    \centering
    \caption{Matching results using the Unscented Transform’s Mahalanobis Distance threshold for Dataset 1 (three-flower cluster) matched against itself. Estimated noise at 0.01 meters}
    \renewcommand{\arraystretch}{1.5} % Increases row height
        \begin{tabular}{|c|c|}
            \hline
            \textbf{Comparison} & \textbf{Correct Match} \\
            \hline
            Data1-Data1 & 4950/4954 \\
            \hline
            Data2-Data1 & 13/277 \\
            \hline
            Data3-Data3 & 1195/2437 \\
            \hline
            Data4-Data3 & 30/1005 \\
            \hline
        \end{tabular}
    \label{tab:data1}
\end{table}

To evaluate the effectiveness of matching a flower cluster across different frames and translations, the results show matches at the corresponding index and subsequent indexes reporting as many additional matches as there are frames.

Table \ref{tab:data1} suggests that the Unscented Transform effectively sets the tolerances for matching. This is supported by the correctly matched frames of a dataset to itself. While the level of perturbations does affect it, the results still indicate the effectiveness of the application of the unscented transform. This is supported by the fact that clusters of the 4954 frames correctly matched 4950 of the frames in matching Dataset 1 to itself under moderate perturbations. Only 4 of the frames failed to match at the index, indicating minimal error. While under intense perturbations, 1195 frames of 2437 were still correctly matched. The results validate the use of the Unscented Transform as a reliable method for establishing a matching tolerance to consistently re-identify clusters.

To assess the uniqueness of the descriptors for matching within the generated tolerances, two different cluster datasets were used. The datasets contained a unique cluster with the same number of flowers. Dataset 1 served as the reference flower cluster for matching, while Dataset 2 contained the flower cluster observed by the robot. Similarly, for the case of a four-flower cluster is observed that Dataset 3 is the reference and Dataset 4 is the observed frames. Table \ref{tab:data1} presents the results of applying the Unscented Transform tolerances from Dataset 1 to Dataset 2 and from Dataset 3 to Dataset 4.

\begin{figure}[htb!]
    \centering
    \includegraphics[width=1\linewidth]{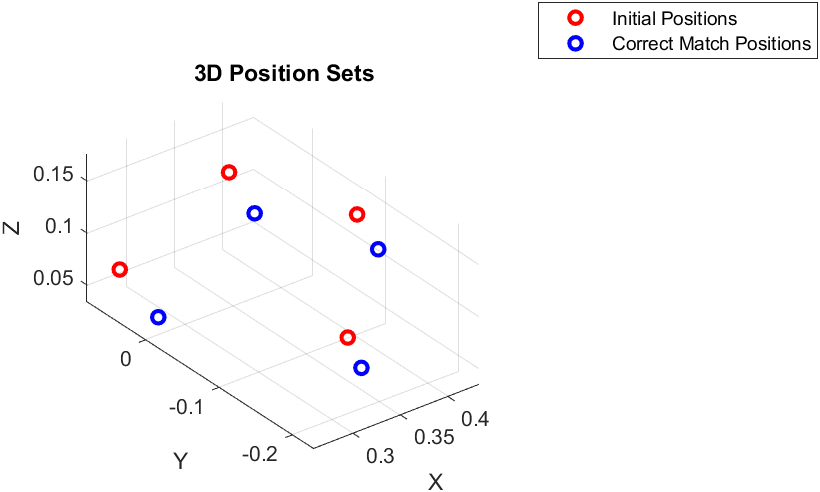}
    \caption{Cluster 3 to Cluster 3 case where it matched}
    \label{fig:match33}
\end{figure}

\begin{figure}[htb!]
    \centering
    \includegraphics[width=1\linewidth]{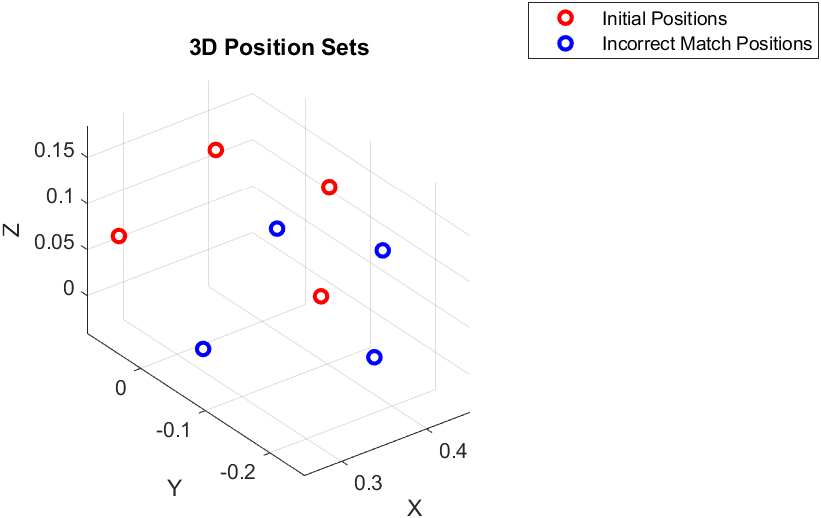}
    \caption{Cluster 3 to Cluster 3 case where it didn't match}
    \label{fig:nonmatch33}
\end{figure}

\begin{figure}[htb!]
    \centering
    \includegraphics[width=1\linewidth]{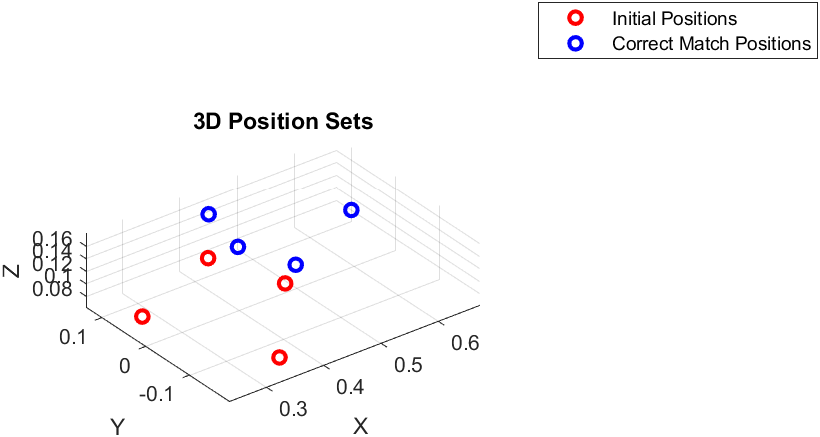}
    \caption{Cluster 4 to Cluster 3 case where it matched}
    \label{fig:match43}
\end{figure}

\begin{figure}[htb!]
    \centering
    \includegraphics[width=1\linewidth]{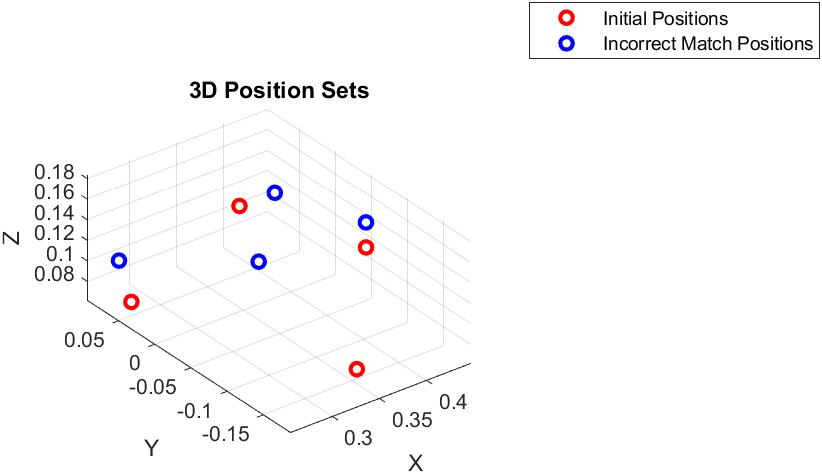}
    \caption{Cluster 4 to Cluster 3 case where it didn't match}
    \label{fig:nonmatch43}
\end{figure}

Only 13 out of 277 frames in Dataset 2 matched the tolerance criteria established by the Unscented Transform for the descriptors of the cluster in Dataset 1. Additionally, for each of these 13 matched frames, the same descriptor also matched 276 other frames within Dataset 2. This suggests that, within Dataset 2, the descriptor tolerances are not significantly unique enough to differentiate the matched clusters from the remaining frames, indicating a high similarity among the flower clusters in Dataset 2 when compared to itself. The results of the four flower clusters matching of Dataset 4 to Dataset 3 further reinforce the robustness.

Fig.s~\ref{fig:match33} -~\ref{fig:nonmatch43} present the cases of matching and non-matching flower clusters observed. The cluster matching approach is robust to translation and rotation, thanks to its consistent descriptors. However, significant differences in cluster formation will result in a non-match. The significant differences in cluster formations are reflected in the descriptors.

\section{Conclusion and Future Works}
\subsection{Conclusion}

In this paper, the novel approach of using the Unscented Transform for cluster descriptor matching tolerance estimation is presented as a reliable method in re-identifying flower clusters with growth and movement. The method presented utilizes a descriptor that describes the points relative to the cluster's centroid. Utilizing the Unscented Transform to handle the nonlinear transformations of the descriptor, a tolerance can be determined for cluster re-identification. The experiments show that the computed covariance and mean are close to those of a Monte Carlo simulation. Additionally, the test with real datasets allowed the successful re-identification of the cluster with minimal errors. This suggests that the approach is robust and reliable under different conditions involved with the precision agricultural processes and the plant's natural growth. This work plays an important role in precision agriculture applications in reducing redundant processes and requiring continuous plant monitoring.

\subsection{Future Works}

Enhancing plant monitoring and re-identification accuracy is crucial due to challenges such as plant similarity, temporal changes, and occlusions. Future work should focus on improving cluster identification (e.g., branch-based grouping, K-means for consistency) to account for growth. YOLO models are a strong real-time alternative to Grounding DINO for lower computational cost. Descriptor refinement can reduce false positives but risks noise sensitivity. These advancements will boost precision agriculture and automation. 

\addtolength{\textheight}{-12cm}   % This command serves to balance the column lengths

%%%%%%%%%%%%%%%%%%%%%%%%%%%%%%%%%%%%%%%%%%%%%%%%%%%%%%%%%%%%%%%%%%%%%%%%%%%%%%%%

\section*{ACKNOWLEDGMENT}

The authors acknowledge Uthman Olawoye for his guidance on the Unscented Transform and the Interactive Robotics Lab for providing the artificial plant. Additionally, the authors acknowledge the use of ChatGPT solely for grammar revisions.

\bibliographystyle{ieeetr}
\bibliography{references}

\end{document}